\documentclass[letterpaper]{mandc2019}
\usepackage{tabls}
\usepackage{cites}
\usepackage{epsf}
\usepackage{appendix}
\usepackage{ragged2e}
\usepackage[top=1in, bottom=1.in, left=1.in, right=1.in]{geometry}
\usepackage{enumitem}
\setlist[itemize]{leftmargin=*}
\usepackage{caption}
\usepackage[capitalize]{cleveref}
\usepackage{amsbsy}
\usepackage{amsmath}
\usepackage{gensymb}
\usepackage{color}
\crefalias{subequation}{equation}
\crefformat{pluraleq}{Eqs.~(#2#1#3)}
\Crefformat{pluraleq}{Equations~(#2#1#3)}

\title{SOLVER RECOMMENDATION FOR TRANSPORT PROBLEMS IN SLABS \\ 
USING MACHINE LEARNING}
\author{%
  \textbf{Jinzhao Chen$^1$, Japan K. Patel$^1$, and Richard Vasques$^1$} \\ 
  $^1$The Ohio State University, Department of Mechanical and Aerospace Engineering\\
201 W.~19$^{\text{th}}$ Avenue, Columbus, OH 43210 \vspace{5pt}\\ 
  \\
  \url{chen.7163@osu.edu}, \url{patel.3545@osu.edu}, \url{vasques.4@osu.edu}
}
\newcommand{\authorHead}      
           {J.~Chen, J.K.~Patel, R.~Vasques}  
\newcommand{\shortTitle}      
           {Solver Recommendation for Transport Problems in Slabs using Machine Learning}  
\begin{document}
\maketitle
\justify 

\begin{abstract}
The use of machine learning algorithms to address classification problems is on the rise in many research areas. 
The current study is aimed at testing the potential of using such algorithms to auto-select the best solvers for transport problems in uniform slabs. 
Three solvers are used in this work: Richardson, diffusion synthetic acceleration, and nonlinear diffusion acceleration. 
Three parameters are manipulated to create different transport problem scenarios. 
Five machine learning algorithms are applied: linear discriminant analysis, K-nearest neighbors, support vector machine, random forest, and neural networks. 
We present and analyze the results of these algorithms for the test problems, showing that random forest and K-nearest neighbors are potentially the best suited candidates for this type of classification problem.
\end{abstract}
\keywords{machine learning, transport solvers, classification algorithms, random forest, neural nets}
 
\section{INTRODUCTION}\label{sec1} 

Several problems in nuclear sciences and engineering involve the transport of particles in media with a wide variety of physical properties.
The accuracy of the chosen transport model and the efficiency of the corresponding solver/preconditioner pair depend on these physical properties.
Numerous studies have been carried out in the computer science and mathematics communities addressing the use of machine learning algorithms for choosing optimal solvers for linear systems \cite{bhowmick,xu,salsa,rlsparse,gmresmri}.
None of these works, however, specifically target particle transport problems, and the authors are unaware of any such study addressing transport solvers.

We show that recommender systems can be developed to choose the best available transport solver using existing machine learning algorithms.
In this preliminary work, we consider two different ways to define the ``best" solver among the available options: 1) total number of transport sweeps, and 2) computation runtime to achieve a prescribed tolerance.
We study transport in uniform slabs, where we examine three numerical solvers: Richardson Iteration \cite{hy}, diffusion synthetic acceleration \cite{alcouffeDSA}, and nonlinear diffusion acceleration \cite{smithNDA}. 
Each of the three solvers uses the finite difference method \cite{lm} for discretization.
We employ five established classification algorithms for our analysis: (i) linear discriminant analysis \cite{mclachlanLDA}; (ii) K-nearest neighbors \cite{altmanKNN}; (iii) support vector machine \cite{benhurSVM}; (iv) neural networks \cite{zhangNN};  and (v) random forest \cite{RandomForestHo,breimanRF}.

The remainder of the paper is organized as follows.
In \cref{sec2}, we briefly describe the machine learning algorithms we use.
We introduce our test problems in \cref{sec3} and present our results and analysis in \cref{sec4}.
Finally, we discuss implications of our findings in \cref{sec5}.

\section{MACHINE LEARNING METHODS} 
\label{sec2}

Recommenders use machine learning algorithms to analyze sample data with relevant parameters, learning to predict (recommend) the desired unknown information. 
We establish our problem as one of supervised classification; in order to solve this classification problem, we must train a classifier.
This type of algorithm can learn hidden patterns from properly labeled data and predict future instances.
In this Section we briefly review the classification algorithms used in this study, as well as the evaluation methods to compare their performances.
The analysis of their applicability for our desired goals is given in \cref{sec4}.
A thorough review and comparison of the algorithms described here can be found in \cite{MLReview}. 

\subsection{Linear Discriminant Analysis}

Linear discriminant analysis (LDA) is a well-known supervised learning method for classification. 
LDA tries to find a linear combination of features (independent variables) that classify two or more classes of events.
LDA assumes that the feature variables are continuous and drawn from a multivariate Gaussian distribution $N(\mu_k, \Sigma)$, where $\mu_k$ is a class-specific mean vector, and $\Sigma$ is a covariance matrix that is common to all $k$ classes.
Based on Bayes' theorem, the LDA classifier can be expressed as follows:
\begin{equation}
\label{lda}
\delta_k(x) = x^T\Sigma^{-1}\mu_k - \frac{1}{2} \mu_k^T \Sigma - \mu_k + \log(\pi_k),
\end{equation} 
where $\delta_k(x)$ is the discriminant score for the observation $x$, and $\pi_k$ is the prior probability that an observation belongs to class $k$. 

The LDA classifier assigns an observation to the $k^{th}$ class when the discriminant score is highest.
The limitation of LDA is that it makes strong assumptions of multivariate normality of feature variables with equal covariance matrixes; this is rare in practice.
A detailed explanation of this method can be found in \cite{mclachlanLDA}.

\subsection{K-Nearest Neighbors}

The K-nearest neighbors (KNN) algorithm is another simple method for discovering underlying patterns in a dataset. 
The KNN algorithm uses data points with known labels to learn how to classify unlabeled data points based on the similarity of these points to the labeled ones.
In this algorithm, data points near to each other are called neighbors. 
KNN is based on the the idea that similar cases with the same class labels are near each other. 
Thus, the distance between two cases is a measure of dissimilarity of two data points. 
The general procedure is as follows:\vspace{-15pt}
\begin{enumerate}
\item Pick a value for number of neighbors, $K$. \vspace{-7pt}
\item Calculate the distance of the new data point from all points in the labeled data set.\vspace{-7pt}
\item Search for $K$ observations in the training data that are nearest to the measurement of the unknown data point. \vspace{-7pt}
\item Predict the response of the unlabeled data point using most popular response value from the $K$ nearest neighbors.\vspace{-7pt}
\end{enumerate}\vspace{-10pt}
The data set is normalized to get accurate dissimilarity measure.
It is important to notice that a high value of $K$ produces an overly generalized model, while an inappropriately low value of $K$ produces an extremely crude model.
A detailed explanation of this algorithm can be found in \cite{altmanKNN}.

\subsection{Support Vector Machine}

A support vector machine (SVM) is a supervised algorithm that can classify cases by finding a separator. 
SVM works by mapping data to a high-dimensional feature space so that data points can be categorized, even when the data is not otherwise linearly separable. 
Then, a separator is estimated for the data. 
The data should be transformed in such a way that a separator could be drawn as a hyperplane. 
One reasonable choice as the best hyperplane is the one that represents the largest separation (margin) between the classes. 
Finding the optimized hyperplane can be formalized using (for example) a linear equation ($y = ax + b$) in the two-dimensional feature space. 
The hyperplane is learned from training data using an optimization procedure that maximizes the margin. 

The two main advantages of SVM are i) accuracy in high-dimensional spaces, and ii) the use of a subset of training points in the decision function (called support vectors). 
The disadvantages include the fact that the algorithm is prone to over-fitting if the number of features is much greater than the number of samples. 
Also, SVMs do not directly provide probability estimates, which are desirable in most classification problems. 
Finally, SVMs are not very efficient computationally when working with large datasets. 
More details can be found in \cite{benhurSVM}.

\subsection{Neural Networks}

Neural networks, or neural nets (NNET), have been applied in classification problems as an alternative to the conventional classifiers.
NNET are used for classification tasks where an object can fall into one of at least two different categories.
Unlike other networks, a neural network is highly structured and comes in layers: the first layer is the input layer; the final layer is the output layer; and all other layers are referred to as hidden layers.
First, a set of inputs is passed to the first hidden layer.
Then, the activations from that layer are passed to the next one.
The process continues until the output layer is reached, where the results of the classification are determined by the scores at each node.
This process happens for each set of inputs.
In other words, a NNET can be viewed as the result of spinning classifiers together in a layered web, with each node in the hidden and output layers having its own classifier. 
For more details on the uses of NNET for classification, we refer the reader to \cite{zhangNN}.

\subsection{Random Forests and Decision Trees}

A random forest (RF) is, essentially, a collection of many decision trees that are created from random samples in the dataset.
Decision trees are built by splitting the training set into distinct nodes, where one node contains all of (or most of) one category of the data.
The construction is done by finding attributes that return the highest information gain, which is the information that can increase the level of certainty after splitting.
The amount of information disorder, or the amount of randomness in the data, is called entropy.
In decision trees, the goal is to look for trees that have the smallest entropy in their nodes.

RF algorithms work roughly as follows.
First, the number of trees that will be built is decided.
Next,  each tree is built upon a bootstrapped sample of data of same size; that is, data points are randomly drawn from the training data with replacement.
Then, for each tree, a random subset of features is selected.
Finally, the feature that gives the best split is used to split the node iteratively.
This whole process is repeated until all trees are built.
The prediction of a RF is the aggregated results of all the decision trees. 

There are two sources of randomness in RF: 1) randomness in the data due to bootstrapping, and 2) randomness in the split of the features.
Randomness is important in RF because it allows for distinct, different trees that are based off of different data.
RF algorithms are very reliable estimators, providing consistently accurate results.
For a more detailed explanation of RF algorithms, see \cite{RandomForestHo,breimanRF}.

\subsection{Evaluation Methods}

The performance of each classifier can be estimated using a series of different statistical metrics.
In this study, we use accuracy (the overall percentage that a classifier makes a correct prediction) and Cohen's kappa statistic \cite{kappa}. 
The Cohen's kappa is calculated as
\begin{equation}
\label{kappa}
\kappa = \frac{p_a - p_e}{1 - p_e},
\end{equation}
where $p_a$ is accuracy, and $p_e$ is the percentage of correct predictions by chance.
The larger the $\kappa$ is, the better the classifier is.
kappa statistic represents a fairer measurement than accuracy alone, as it ``corrects" the estimated performance by taking chance into account.

In addition, the reliability of these metrics can be ensured by applying the $K$-fold cross-validation scheme \cite{kfoldCV}.
The training set is randomly divided into $K$ disjoint sets of equal size, with each part having roughly the same class distribution.
The classifier is trained $K$ times, each time with a different set held out as a test set.
The estimated performance metrics are the average of all the tests.
In this work, we use 4-fold cross-validation with 25 repetitions to measure the performance of the classification algorithms.

\section{TEST PROBLEMS}\label{sec3}

We use classification algorithms to recommend the best solver (with respect to number of transport sweeps or computation runtime) for the following one-speed, steady-state, slab geometry transport equation with isotropic scattering and uniform isotropic source:
\begin{subequations}\label[pluraleq]{eq3}
\begin{align}
\label{transport}
&\mu \frac{\partial}{\partial x} \psi(x, \mu) + \sigma_t\psi(x,\mu) = \frac{\sigma_s}{2}\phi(x) + \frac{Q}{2},\\
&\psi(x_l,\mu) = 0, \quad \mu>0,\quad \psi(x_r,\mu) = 0, \quad \mu<0,
\end{align}
\end{subequations}
where $\psi$ is the angular flux at position $x$ along angular cosine $\mu$, $\sigma_t$ and $\sigma_s$ are the macroscopic total and scattering cross-sections, $Q$ is the internal source, $\phi = \int_{-1}^{1} d \mu \psi$ is the scalar flux, $x_l$ is the position of the left boundary and $x_r$ is the position of the right boundary.
We set the slab thickness to $10$ cm, the macroscopic total cross-section to unity, and the source strength to $6$ $\frac{neutrons}{cm^3 s}$.
We use three different methods to solve this equation.
While all of these methods are well-known, we discuss them here briefly for the sake of completeness.  

\textit{Richardson (source) iteration} is the simplest of all iterative methods.
We simply iterate on the source and keep iterating until convergence (for convenience, we drop the notation for $x$ and $\mu$ dependence henceforth):
\begin{equation}
\label{richardson}
\mu \frac{\partial}{\partial x} \psi^{k+1} + \sigma_t\psi^{k+1} = \frac{\sigma_s}{2}\phi^{k} + \frac{Q}{2},
\end{equation}
where $k$ is the iteration index.
The convergence criterion is set such that the relative error $\epsilon$ in the $L_2$ norm, given by
\begin{equation}
\label{conv_criteria}
\epsilon = \frac{||\phi^{k+1} - \phi^{k}||}{||\phi^{k+1}||},
\end{equation}
must reduce to a value below a prescribed tolerance.
In this work, we set our tolerance to $10^{-5}$ for all three solvers.

We discretize the transport equation using diamond difference in space and discrete ordinates in angle (we drop the iteration index $k$ from our notation for convenience):
\begin{subequations}
\begin{align}
\label{rtdisc}
&\mu_n \frac{\psi_{i+ \frac{1}{2}, n} -\psi_{i- \frac{1}{2}, n}}{\Delta x} + \sigma_{t}\psi_{i,n} = \frac{\sigma_{s}}{2}\phi_{i} + \frac{Q_i}{2},\\
\label{closure1}
&\psi_{i, n} = \frac{\psi_{i+ \frac{1}{2}, n}+\psi_{i- \frac{1}{2}, n}}{2}.
\end{align}
\end{subequations}
Here, $i$ and $n$ are the spatial and angular discretization indexes, respectively. 

\textit{Diffusion synthetic acceleration} (DSA) essentially follows a predict-correct-iterate scheme.
The predictor step is a transport sweep:
\begin{equation}
\label{richardson}
\mu \frac{\partial}{\partial x} \psi^{k+\frac{1}{2}} + \sigma_t\psi^{k+\frac{1}{2}} = \frac{\sigma_s}{2}\phi^{k} + \frac{Q}{2}.
\end{equation}
The predicted angular flux value, $\psi^{k+\frac{1}{2}}$, is then corrected using an approximate error equation.
For DSA, this is a diffusion equation:
\begin{equation}
\label{diff}
-\frac{1}{3\sigma_t} \frac{d^2}{d x^2} f^{k+1} + \sigma_a f^{k+1} = \frac{\sigma_s}{2} (\phi^{k + \frac{1}{2}} - \phi^k),
\end{equation}
where adding the correction $f^{k+1}$ to the predicted value of flux from the transport sweep returns the next iterate of the angular flux:
\begin{equation}
\label{correct}
\psi^{k+1} = \psi^{k+\frac{1}{2}} + f^{k+1}.
\end{equation}
We iterate until the convergence criterion is met.
We discretize the predictor step as in \cref{rtdisc}.
The standard cell-centered finite difference scheme, however, does not return the desired spectral radius.
Special regularization is introduced in the discretization of \cref{diff} to achieve the theoretical convergence rate \cite{wang}:
\begin{equation}
\label{diffdisc}
\frac{-1}{3\sigma_t} \left(\frac{f_{i-1} - 2f_{i} + f_{i+1}}{\Delta x^2}\right) + \sigma_a \left(\frac{f_{i-1} + 2f_i + f_{i+1}}{4}\right) = \frac{\sigma_s}{2} (\phi_i^{k+\frac{1}{2}} - \phi_i^k), 
\end{equation}
where $\Delta x$ is the meshsize.
The correction is also regularized \cite{wang}:
\begin{equation}
\label{correct}
\psi_i^{k+1} = \psi_i^{k+\frac{1}{2}} + \frac{f_{i-1}^{k+1}+2f_{i}^{k+1}+f_{i+1}^{k+1}}{4}.
\end{equation}
\textit{Nonlinear diffusion acceleration} (NDA) introduces a discretely consistent low-order (LO) equation  \cite{smithNDA} and solves the following system of equations simultaneously:
\begin{subequations}
\begin{align}
\label{ho1}
&\mu\frac{\partial \psi^{HO}}{\partial x} + \sigma_t \psi^{HO} =  \frac{\sigma_{s}(x)}{2} \phi^{LO} + \frac{Q}{2},\\
\label{lo}
&\frac{d}{dx} \left(-\frac{1}{3\sigma_t}\frac{d \phi_{LO}}{d x} + \hat{D}\phi^{LO}\right) + \sigma_a \phi^{LO} = Q,\\
\label{c1}
&\hat{D} = \frac{J^{HO} + \frac{1}{3\sigma_t}\frac{d \phi^{HO}}{d x}}{\phi^{HO}},\\
\label{c2}
&J^{HO} = \int_{-1}^1 \mu \psi^{HO} d\mu.
\end{align}
\end{subequations}
Here, $J$ is the current, $D = \frac{1}{3 \sigma_t}$ is the diffusion coefficient, and $\hat{D}$ is the consistency parameter. 
The high-order (HO) transport equation (\ref{ho1}) couples with a diffusion-like LO equation (\ref{lo}) via the closure relations given by \cref{c1,c2}. 
The HO equation is discretized just as \cref{rtdisc}, while the LO equation employs the standard central finite difference for spatial discretizations \cite{knollJFNK}:
\begin{equation}
\label{lodisc}
-\frac{1}{3\sigma_t} \frac{\phi_{i+1} -2\phi_{i}+ \phi_{i+1}}{\Delta x^2} +\frac{\hat{D}_{i+\frac{1}{2}}}{2 \Delta x}(\phi_{i+1} + \phi_{i}) - \frac{\hat{D}_{i-\frac{1}{2}}}{2 \Delta x}(\phi_{i} + \phi_{i-1}) + \sigma_a \phi_{i}  = Q_i.
\end{equation}
The fractional indices represent cell-edge terms while integer indices represent cell centers. 


\section{DATA ANALYSIS}\label{sec4}

We use the recommender systems described in \cref{sec2} to predict (recommend) the most well-suited of the methods described in \cref{sec3} to solve \cref{eq3}.
Specifically, these algorithms will classify the three solvers (DSA, NDA, and Richardson) based on which solver performed the best as a function of three features: the spatial discretization size, the $S_N$ order, and the scattering ratio $c=\sigma_s/\sigma_t$.
These features are given by:
\vspace{-10pt}
\begin{enumerate}
\item Five different choices of S$_N$ orders: $N = 2, 4, 8, 16, 32$.\vspace{-5pt}
\item Nine choices of spatial discretization refinements (cells): $2^i$, where $i = 2,3,4,...,10$.\vspace{-5pt}
\item One hundred and one different choices of scattering ratios: $c=0, 0.01, 0.02, ..., 0.98, 0.99, 1.0$.\vspace{-10pt}
\end{enumerate}
We solved \cref{eq3} for each of the $4,545$ combinations of these features using all three solvers: Richardson, NDA, and DSA.
All computations were performed in MATLAB (version R2018a) in a PC with Windows 64-bit OS, 16GB RAM, and 3.50GHz Intel Processor. 
Then, we compared the number of iterations and runtimes and created an explicit dataset identifying the combinations in which each of the solvers outperforms the others.
\begin{figure}[h]
\centering
  \includegraphics[scale=0.35]{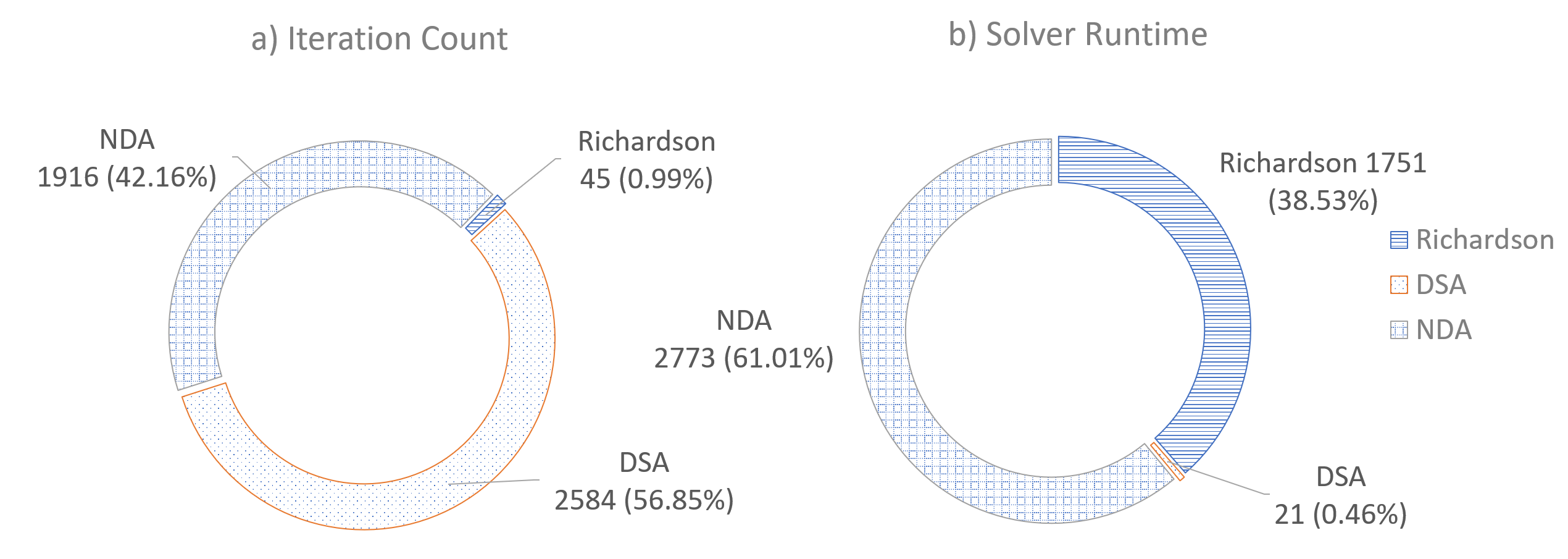}
   \caption{\small Relative proportion of the best solvers in the dataset of 4,545 problems}
  \label{fig:fig1}
\end{figure}

The resulting distribution of best solvers identified in our dataset is shown in \cref{fig:fig1}.
We observe that the classes are unbalanced to some degree in both scenarios.
Richardson accounts for less than 1\% of the best solvers with respect to iteration count and DSA is the best solver concerning runtime in more than 68\% of the cases.
This imbalance results in bias in accuracy metric but not in the kappa statistic, since kappa statistic takes care of the probability of chance.
As a result, we perform multi-class classification tasks on choosing the best solver based on three problem features.

\subsection{Numerical Results}

We fit various classification models with our experimental dataset, treating ``best solver" as a function of three transport problem properties (S$_N$ order, number of cells, and scattering ratio).
We applied 4-fold cross-validation for 25 times to increase robustness of our measurements.
All the analysis is performed in R (version 3.5.0) using the caret package \cite{rCaret}.
For simplicity, we used default values for hyper-parameters for all models.
Results are summarized in \cref{table:tab1,table:tab2} for classifying best solvers by iteration count and by runtime, respectively.
In the tables, the average classification accuracy and the average kappa statistic (both with standard deviation) are shown, along with the modeling efficiency (i.e. computational duration) measured in seconds.
The five models are ranked by their average accuracy.

We found that, when defining best solver based on iteration count, the RF model achieved the highest average accuracy of 0.972 and the highest kappa value of 0.946, followed by the KNN model, NNET, SVM, and LDA.
In addition, we found that SVM is the least efficient model with 451 seconds used in the total cross-validation process, while LDA is the fastest model with less than 3 seconds (\cref{table:tab1}).
The results are similar when we define best solver based upon runtime, with RF performing the best in both accuracy and kappa (0.960 and 0.915).
\begin{table}[ht]
\caption{Classifying Best Solvers by Iteration Count 
} 
\centering 
\begin{tabular}{c c c c c} 
\hline\hline 
Rank & ML Algorithms & Accuracy & Kappa Measure & Modeling Efficiency \\ [0.5ex] 
\hline 
1 & RF & 0.972 (0.005) & 0.946 (0.010) & 181.37 s \\ 
2 & KNN & 0.919 (0.007) & 0.840 (0.014) & 10.88 s \\
3 & NNET & 0.765 (0.021) & 0.535 (0.041) & 324.97 s \\
4 & SVM & 0.686 (0.014) & 0.374 (0.028) & 451.74 s \\
5 & LDA & 0.612 (0.015) & 0.231 (0.029) & 2.39 s \\ [1ex] 
\hline 
\end{tabular}
\label{table:tab1} 
\end{table}
\begin{table}[ht]
\caption{Classifying Best Solvers by Runtime
} 
\centering 
\begin{tabular}{c c c c c} 
\hline\hline 
Rank & ML Algorithms & Accuracy & Kappa Measure & Modeling Efficiency \\ [0.5ex] 
\hline 
1 & RF & 0.960 (0.005) & 0.915 (0.011) & 151.11 s \\ 
2 & KNN & 0.952 (0.006) & 0.899 (0.012) & 10.59 s \\
3 & NNET & 0.929 (0.007) & 0.851 (0.020) & 391.41 s \\
4 & SVM & 0.924 (0.007) & 0.837 (0.016) & 163.79 s \\
5 & LDA & 0.876 (0.007) & 0.721 (0.017) & 2.25 s \\ [1ex] 
\hline 
\end{tabular}
\label{table:tab2} 
\end{table}

\begin{figure}[h]
\centering
  \includegraphics[scale=0.57]{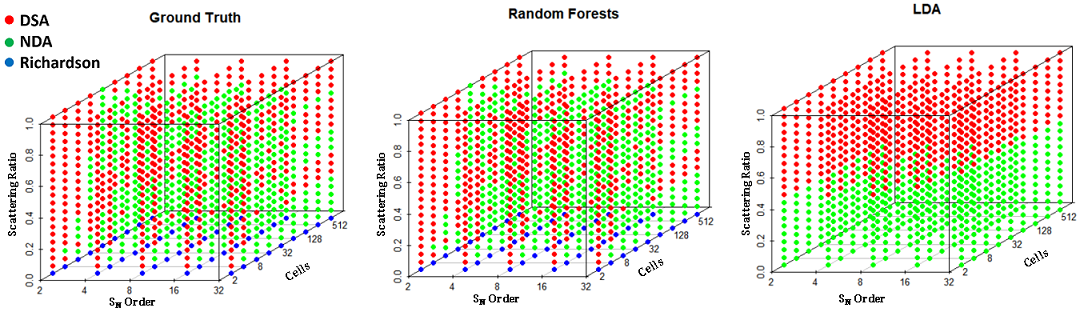}
   \caption{\small Best Solver Distribution in Combination of Features (by Iteration)\vspace{30pt}}
  \label{fig:fig3}
   \includegraphics[scale=0.57]{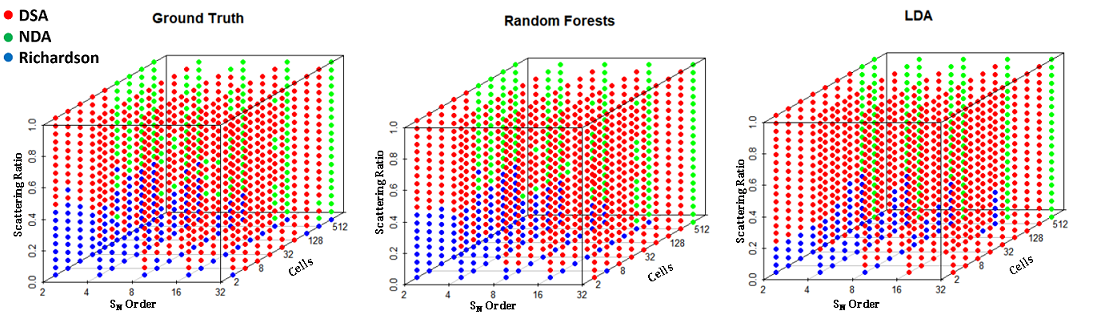}
   \caption{\small Best Solver Distribution in Combination of Features (by Runtime)}
  \label{fig:fig4}
\end{figure}

A comparison of prediction performance factored into the feature space between the best model (RF) and the worst model (LDA) is shown in \cref{fig:fig3} (by iteration count) and \cref{fig:fig4} (by runtime).
Regarding best solver with respect to number of iterations, the RF model predicts 45 cases of Richardson with precision of 100\%, 2193 cases of NDA with precision of 97\%, and 2307 cases of DSA with precision of 98\%.
On the other hand, the LDA model predicts none of the cases of Richardson, 2144 cases of NDA with precision of 59.7\%, and 2401 cases of DSA with precision of 62.8\%.

In the case of best solver according to runtime, the RF model predicts 597 cases of Richardson with precision of 93.8\%, 814 cases of NDA with precision of 92.3\%, and 3134 cases of DSA with precision of 97.5\%.
The LDA model predicts 424 cases of Richardson with precision of 85.1\%, 717 cases of NDA with precision of 81.6\%, and 3404 cases of DSA with precision of 89.3\%.
In summary, RF performed much better than LDA, especially when the number of cases in each class is dramatically different (such as in the iteration scenario). 

\section{DISCUSSION}\label{sec5}

We have demonstrated the capability of machine learning models in predicting the best solvers of a simple transport slab problem with desired parameter settings.
We consistently verified that RF performed better than other models in terms of both accuracy and kappa measure.
When model efficiency is taken into account, one would be tempted to choose KNN as the recommender model; it is 15 times faster than RF and maintains a comparable accuracy and kappa statistic.
However, RF can automatically give us the relative importance of the features based on Gini impurity \cite{gini}.
In our study, the final RF model shows that the scattering ratio has the highest mean decrease in Gini impurity (976.24), followed by number of cells (739.83), and S$_N$ order (483.79).
We note that a higher mean decrease in Gini impurity indicates a higher variable importance \cite{gini}.
This is important information for feature selection and to enhance future predictive models.

\begin{figure}[h]
\centering
  \includegraphics[scale=0.5]{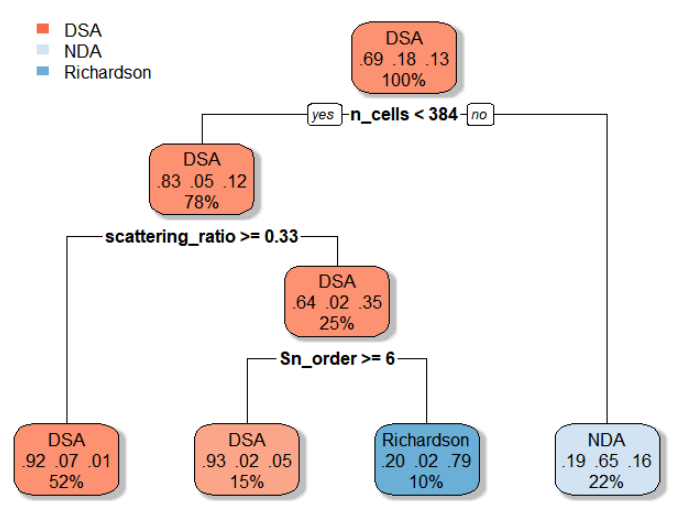}
   \caption{\small Visualization of a decision tree among all the trees in the random forest model.}
  \label{fig:fig5}
\end{figure}

Another advantage of using the RF model is its interpretability.
One can easily examine the logic behind each decision tree that is built within the RF model.
For example, in the second scenario of this study (best solver by runtime), we can visualize \textit{one} of the trees in the ``forest" as shown in \cref{fig:fig5}.
At each node, there is information on 1) the name of the major best solver; 2) proportion of DSA, NDA, and Richardson in each (sub)sample; and 3) the percentage of the sample size at the current node.
There is a rule based on a feature associated with each split of the nodes (e.g., ``number of cells $<$ 384" splits the top node).
One can quickly make sense of the decision logic within the relevant context.

Furthermore, we also observe that the LDA algorithms were the most efficient but performed the worst among all the models.
This is likely due to the fact that LDA is too simple to capture the underlying relationship between the features and the best solver choices.
Also, SVM did not perform very well in our test scenarios.
This is because we only have three variables, and SVM generally works better in high-dimensional problems.

NNET models tend to work better when the underlying pattern between inputs and outputs is nonlinear and complex.
Since NNET performed significantly better in scenario 2 (classifying best solvers by runtime) than in scenario 1 (classifying best solvers by iteration count), we infer that the underlying relation between the three features and the true best solvers is more complex in scenario 2 than scenario 1.
This trait needs further investigation to be better understood.

Finally, our findings imply that some of the existing machine learning algorithms, even with the simplest configuration, are capable of predicting suitable solvers for certain settings of a transport problem with high accuracy.
In particular, we recommend RF and KNN as the best models in such problems.
One can fine-tune the parameters of these models to achieve better results.
The validity of these machine learning models needs to be investigated with more complex transport problems and with a larger number of problem features.
This will be the focus of our future work, in addition to increasing the number of solvers being compared.

\section*{ACKNOWLEDGMENTS}

J.~K.~Patel and R.~Vasques acknowledge support under award number NRC-HQ-84-15-G-0024 from the Nuclear Regulatory Commission.
The statements, findings, conclusions, and recommendations are those of the authors and do not necessarily reflect the view of the U.S. Nuclear Regulatory Commission.
\vspace{-5pt}

 \bibliographystyle{mandc}
 \bibliography{mandc}

%

\end{document}